\documentclass{nextgame2026} 

\usepackage{microtype}
\usepackage{booktabs}
\usepackage{multirow}
\usepackage{commands}
\usepackage{float}
\usepackage{balance}
\usepackage{placeins}
\usepackage{mathtools}
\usepackage[capitalize,noabbrev]{cleveref}
\usepackage[disable,textsize=tiny]{todonotes}
\usepackage[normalem]{ulem}

\title[Towards Learning Representations of Policies in Games]{Towards Learning Representations
of Policies in Two-Player Zero-Sum Imperfect-Information Games}

\optauthor{%
\Name{Kevin Wang$^*$} \Email{kevin\_a\_wang@brown.edu}\\
\Name{Kevin Yang$^*$} \Email{kevin\_c\_yang@brown.edu}\\
\Name{Arjun Prakash} \Email{arjun\_prakash@brown.edu}\\
\Name{Amy Greenwald} \Email{amy\_greenwald@brown.edu}\\
\addr Brown University}

\begin{document}

\insert\footins{\noindent\footnotesize\textsuperscript{*}Equal contribution.\par\vspace{1ex}}

\maketitle
\begin{abstract}%
We investigate the problem of learning useful policy representations (embeddings) in two-player zero-sum imperfect-information games. We make three  contributions: First, we introduce methods of creating datasets of policies for a given game. Second, we propose methods to learn policy representations. Third, we introduce downstream tasks to evaluate the effectiveness of such  representations.

We evaluate each dataset method, embedding method, and downstream task on Kuhn and Leduc Poker.
Although our methods are very basic, we demonstrate that useful behavioral representations are present in the learned embeddings. To our knowledge, this work is among the first to systematically compare self-supervised learning techniques for learning policy representations in games.
Our code is available at \url{https://github.com/VitamintK/ssl-project} for others to extend.

\end{abstract}

\section{Introduction}

In two-player zero-sum imperfect-information games, agents can benefit from reasoning about their opponent's policies, and their own. In general, policies in such games are stochastic.\footnote{In contrast, in single-agent or perfect-information settings, reasoning about individual actions typically suffices -- e.g. lookahead search in chess and go.}

For example, an agent seeking to play the Nash equilibrium at some decision point in a hand of poker may perform lookahead search by imagining that each player plays some policy for a few steps, and then evaluating the resulting public belief state. This depth-limited search is typically done tabularly\footnote{e.g. using counterfactual regret minimization or linear programming}~\cite{sokota_solving_2021, brown_combining_2020}, which is feasible since such a depth-limited policy has a tractable size.
However, in games with larger public belief states, such a policy becomes intractable to enumerate. Thus, to perform a version of such a search, an agent will need to reason with compact representations of policies.

So we want to learn good, compact representations of policies. However, there is little existing research towards this end, especially in the context of two-player zero-sum imperfect-information games. There are few existing methods to learn policy embeddings, or comparisons between methods. In order to compare methods, we also need an extensive set of evaluations, which does not exist.

In this work, we begin an extensive, systematic approach to the problem of learning compact, useful representations of policies, particularly in two-player zero-sum imperfect-information games. We are particularly interested in methods that allow decoding embeddings into policies and in representations that can predict future payoffs.

We do this in three parts:
\begin{enumerate}
    \item We introduce three methods of creating datasets of policies for a given game.~(Section \ref{subsec:policy-population})
    \item We propose several methods of varying complexity for learning representations of policies. We also reimplement an existing method.~(Section \ref{subsec:learning})
    \item We introduce several downstream tasks to evaluate the usefulness of the policy representations, and we use them to evaluate the methods.~(Section \ref{subsec:downstream-tasks})
\end{enumerate}

\section{Preliminaries}
\paragraph{Imperfect Information Games}
An imperfect-information game (IIG) is one in which each player may not know the true state of the world. A two-player zero-sum game is one in which there are two players, and the players' payoffs sum to 0. Formally, an IIG is given by the tuple \cite{rudolph2025reevaluating}:
$
\langle \States, \Actions, \Obs, \InfoSet, \Rewardfunc, \Transfunc, \Obsfunc, \Choicefunc, \tmax \rangle ,
$
where $\States$ is the space of game states, $\Actions$ is the action space, $\Obs$ is the observation space and $\InfoSet = \cup_t(\Obs \times \Actions)^t \times \Obs$ is the space of information sets.
The reward function is given by $\Rewardfunc:\States \times \Actions \to \R$ and $\Transfunc: \States \times \Actions \to \simplex (\States)$ is the transition function. The observation function, which determines which observation is given to the acting player is $\Obsfunc: \States \to \Obs$ and $\Choicefunc: \States \to \{ 1,2\}$ is the choice function which determines the acting player and $\tmax$ is the timestep at which the game ends. Players interact with the game using policies $\policy[i]:\InfoSet \to \simplex(\Actions)$ for $i \in \{1,2\}$ which maps from information set to a distribution over actions.
The objective of player 1 is to maximize expected return:
$
\obj = \E_{\policy} \big[\sum^{\tmax}_{t=0} \Rewardfunc(\state[][t], \action[][t]) \big],
$
where $\state[][t] \in \States$ and $\action[][t] \in \Actions$ are the states and actions at time $t$.

\paragraph{Related Works} To save space, we describe related works in detail in Appendix~\ref{sec:related-works}.




\section{Methods}
\subsection{Methods for Making Datasets of Policies}
\label{subsec:policy-population}

In order to learn and evaluate representations of policies, we need a dataset of policies. We propose three methods of generating such datasets:

In our first method, we simply randomly initialize policy neural networks, producing a diverse population of behaviorally distinct agents. We use a custom initialization for the parameters of the neural networks, as the PyTorch default initialization produces behaviorally homogenous policies~(Appendix~\ref{sec:detail-init}).

Our second method of producing a set of diverse policies for a game is to run the \textbf{Policy Space Response Oracle (PSRO)} algorithm \cite{lanctot2017unified}, a method that iteratively trains best-responses to mixtures of other policies to produce an expanding pool of policies.

Our third method of producing a set of diverse policies for a game is to run a variant of \textbf{neural population learning (NeuPL)}~\cite{liu2022neupl,liu_neural_2024}, which is the same as PSRO, but all policies share the same conditional neural network. After training the network (which is multiple agents), we sample from its latent embedding space to generate new populations of policies.


\subsection{Methods for Learning Embeddings}
\label{subsec:learning}

Here, we present several methods for learning vector embeddings that represent policies. Some methods we present are simple and naive, and others are comparatively more complex. We assume that policies are policy neural networks. In this work, we use policy neural networks which are MLPs with 3 hidden layers of size 256, and we train them (when needed) via PPO~\cite{schulman_proximal_2017}.

\begin{figure}[h]
    \begin{minipage}[t]{0.48\textwidth}
      \centering
      \includegraphics[width=\linewidth]{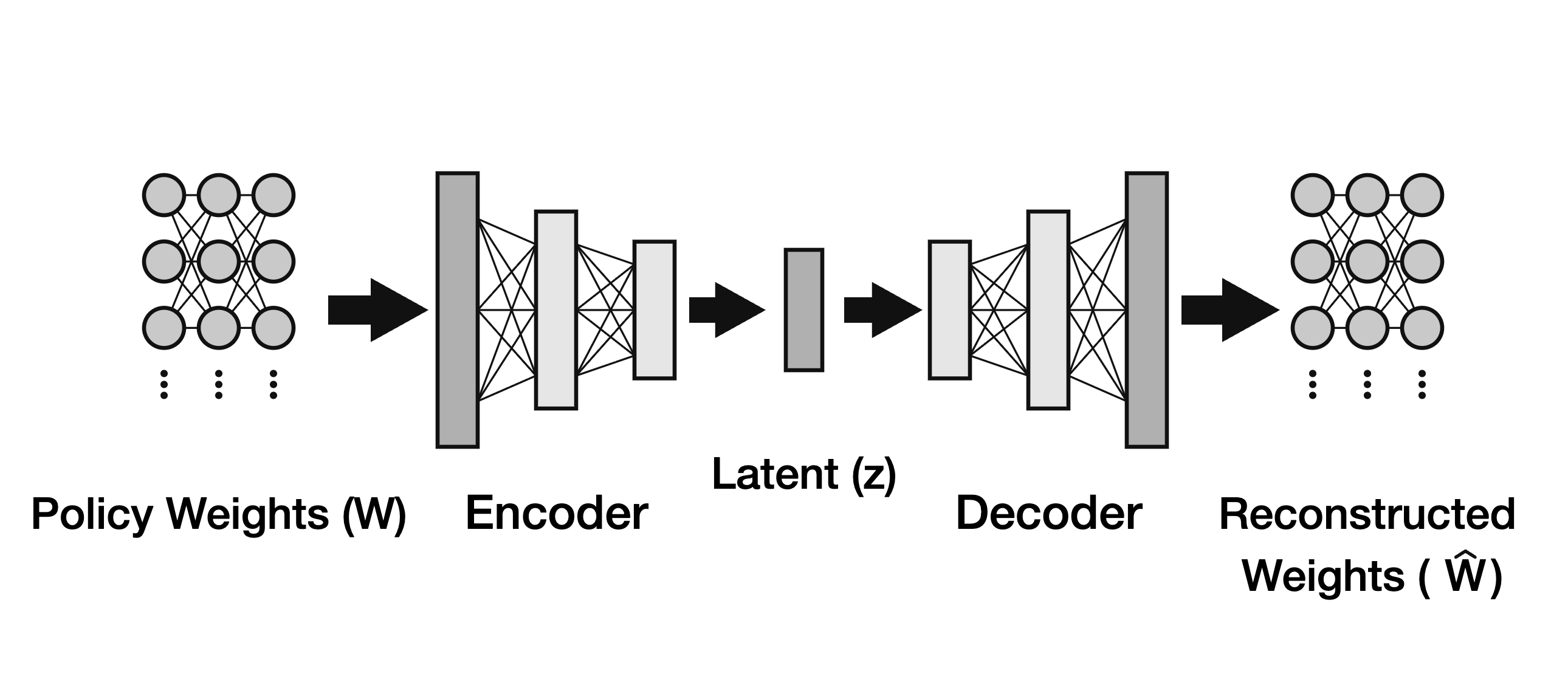}
      \caption{Weight autoencoder}
      \label{fig:wae}
    \end{minipage}
    \hfill
    \begin{minipage}[t]{0.48\textwidth}
      \centering
      \includegraphics[width=\linewidth]{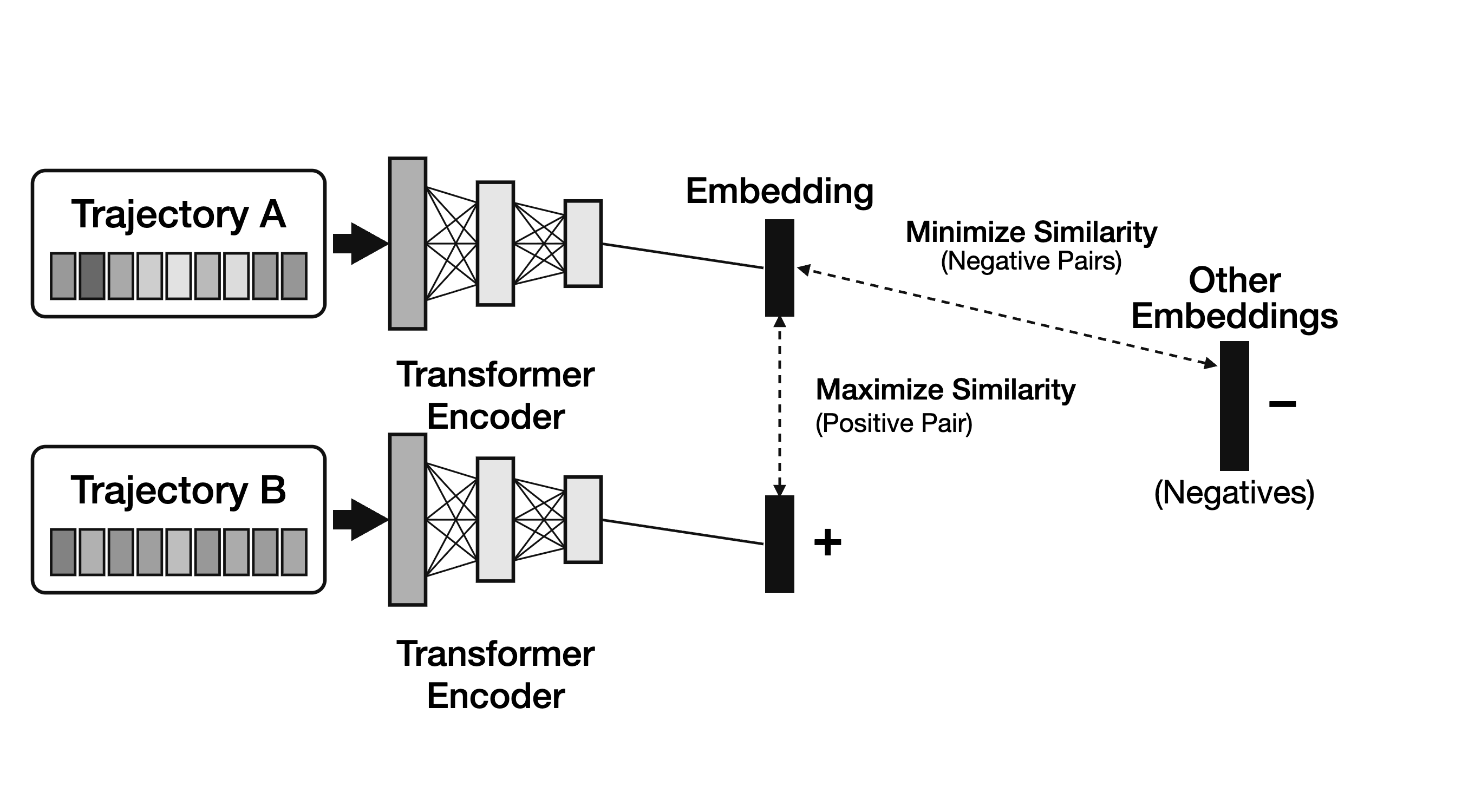}
      \caption{Trajectory encoder}
      \label{fig:te}
    \end{minipage}
\end{figure}

\paragraph{Weight Autoencoder} The first naive method is to autoencode the weights of the policy neural network.\footnote{Here, we colloquially refer to all the parameters of the neural network (both weights and biases) as ``weights''.} The weight autoencoder (Figure~\ref{fig:wae}) takes as input the flattened vector of policy weights and reconstructs the weights from a bottleneck layer. Given an encoder $f_\phi: \R^d \to \R^k$ and decoder $g_\psi: \R^k \to \R^d$, we obtain embeddings $\embed = f_\phi(\params)$ and minimize the reconstruction loss:
$$
\calL_{\text{WAE}} = \| \params - g_\psi(f_\phi(\params)) \|_2^2.
$$



While a policy \emph{is} (in a sense) its parameter vector, making weight autoencoders a natural baseline, we hypothesize this approach will perform poorly: behavior is difficult to predict from raw weights, and the large input/output dimension forces either an intractably large autoencoder or an overly compressed bottleneck.

We note that the decoder is a hypernetwork~\cite{ha2016hypernetworks}. In this work, we simply use an MLP for the autoencoder. However, it is plausible that state-of-the-art hypernetwork architectures may make this type of method more feasible~\cite{chauhan_brief_2024}.

\paragraph{Functional Encoder} The functional encoder\footnote{diagram in Appendix~\ref{sec:baseline-diagrams}} uses the same encoder-decoder architecture as the weight autoencoder, but instead of reconstructing weights directly, it evaluates both the original and reconstructed networks on randomly sampled information states $I \sim \InfoSet$. Let $\hat{\params} = g_\psi(f_\phi(\params))$ denote the reconstructed weights. The loss minimizes the KL divergence between the action distributions:
$$
\calL_{\text{FAE}} = \E_{I \sim \InfoSet} \Big[ D_{\text{KL}}\!\big(\policy[\params](\cdot \mid I) \;\|\; \policy[\hat{\params}](\cdot \mid I)\big) \Big].
$$
In this way, the autoencoder reconstructs the \emph{behavior} of the original network rather than the weights themselves.


\paragraph{Trajectory Encoder}
To capture the behavioral semantics of policies rather than their internal network architecture, we use a trajectory encoder (\Cref{fig:te}) trained with contrastive learning inspired by SimCLR \cite{chen2020simple}. A trajectory $\tau = (o_0, a_0, r_0, \ldots, o_T, a_T, r_T)$ is collected by rolling out a policy against a randomly sampled opponent from the population. An encoder $h_\phi$ maps variable-length trajectories to a fixed-dimensional latent space: $\embed = h_\phi(\tau)$.

For a mini-batch of $N$ policies, we sample two trajectories per policy (against different opponents), yielding $2N$ embeddings. Let $\embed[i]$ and $\embed[j]$ be a positive pair (two trajectories from the same policy). We optimize the normalized temperature-scaled cross-entropy (NT-Xent) loss:
$$
\calL_{\text{NT-Xent}} = -\frac{1}{2N}\sum_{(i,j)} \log \frac{\exp\!\big(\text{sim}(\embed[i], \embed[j]) / t\big)}{\displaystyle\sum_{\substack{k=1 \\ k \neq i}}^{2N} \exp\!\big(\text{sim}(\embed[i], \embed[k]) / t\big)},
$$
where $\text{sim}(\cdot, \cdot)$ denotes cosine similarity and $t$ is a temperature parameter. This objective ensures embeddings are invariant to stochastic environmental noise yet discriminative of strategic intent.

Like the previous methods, this allows us to embed a given policy. In addition, we can learn and create embeddings for arbitrary policies with only access to their trajectories, without knowledge of the neural network internals, and without the requirement that policies are neural networks. Unlike the other methods, this doesn't directly give us a method for turning an embedding into a policy.

\begin{figure}[bth]
    \centering
    \includegraphics[width=0.5\linewidth]{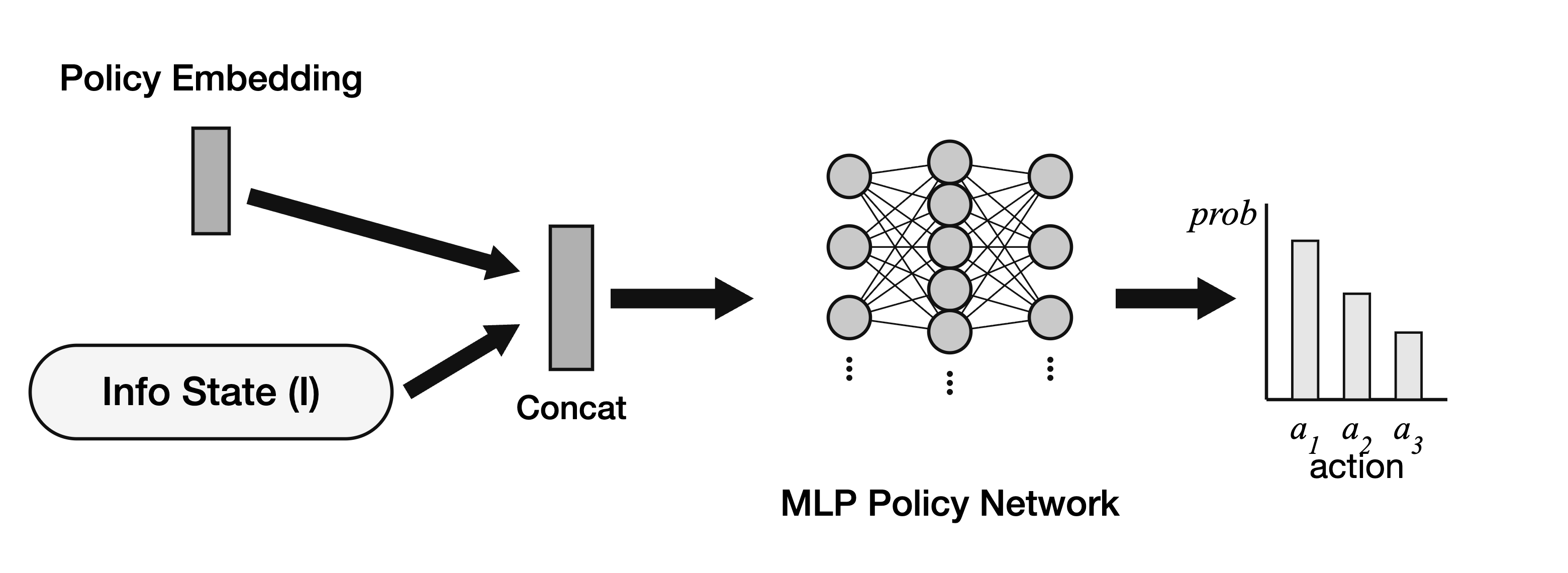}
    \caption{NeuPL-style conditional policy network}
    \label{fig:neupl}
\end{figure}

\paragraph{NeuPL-style Conditional Embeddings}

Training NeuPL induces an embedding space in which any vector can be decoded into a policy via the conditional network (\Cref{fig:neupl}). Unlike the other methods, this generates the population and embeddings jointly rather than post-hoc.

\paragraph{Grover et al.\ Encoder}

Following \citet{grover2018learning}, we implement a hybrid generative-discriminative baseline that learns policy embeddings from episode data. An MLP encoder $f_\theta$ processes individual (observation, action) pairs and averages the resulting vectors across the episode to form the embedding $\embed = \frac{1}{T}\sum_{t=1}^{T} f_\theta(o_t, a_t)$. A conditional policy network $\pi_{\phi,\theta}(\action | o, \embed)$ predicts the agent's actions given an observation and the embedding. The model is trained with a hybrid loss combining an imitation (behavioral cloning) term and a triplet-based agent identification term.
It differs from the trajectory encoder in both architecture (Transformer vs.\ MLP with mean pooling) and training objective (contrastive NT-Xent vs.\ hybrid imitation and triplet identification).

\paragraph{Tabular Baseline}
As a non-learned baseline, we compute each agent's complete action distribution at every information state in the game by querying the policy network. Concatenating these distributions yields a fixed-dimensional vector that fully describes the policy's behavior. This tabular representation requires no training.

\paragraph{Identity Baseline}
We can also test the usefulness of using the weights of the neural network directly, without embedding them to a lower-dimensional representation. For some tasks, this very high-dimensional representation is too expensive to be tractable.

\begin{figure*}[th]
    \centering
    \begin{minipage}{0.38\textwidth}
        \centering
        \includegraphics[width=\linewidth]{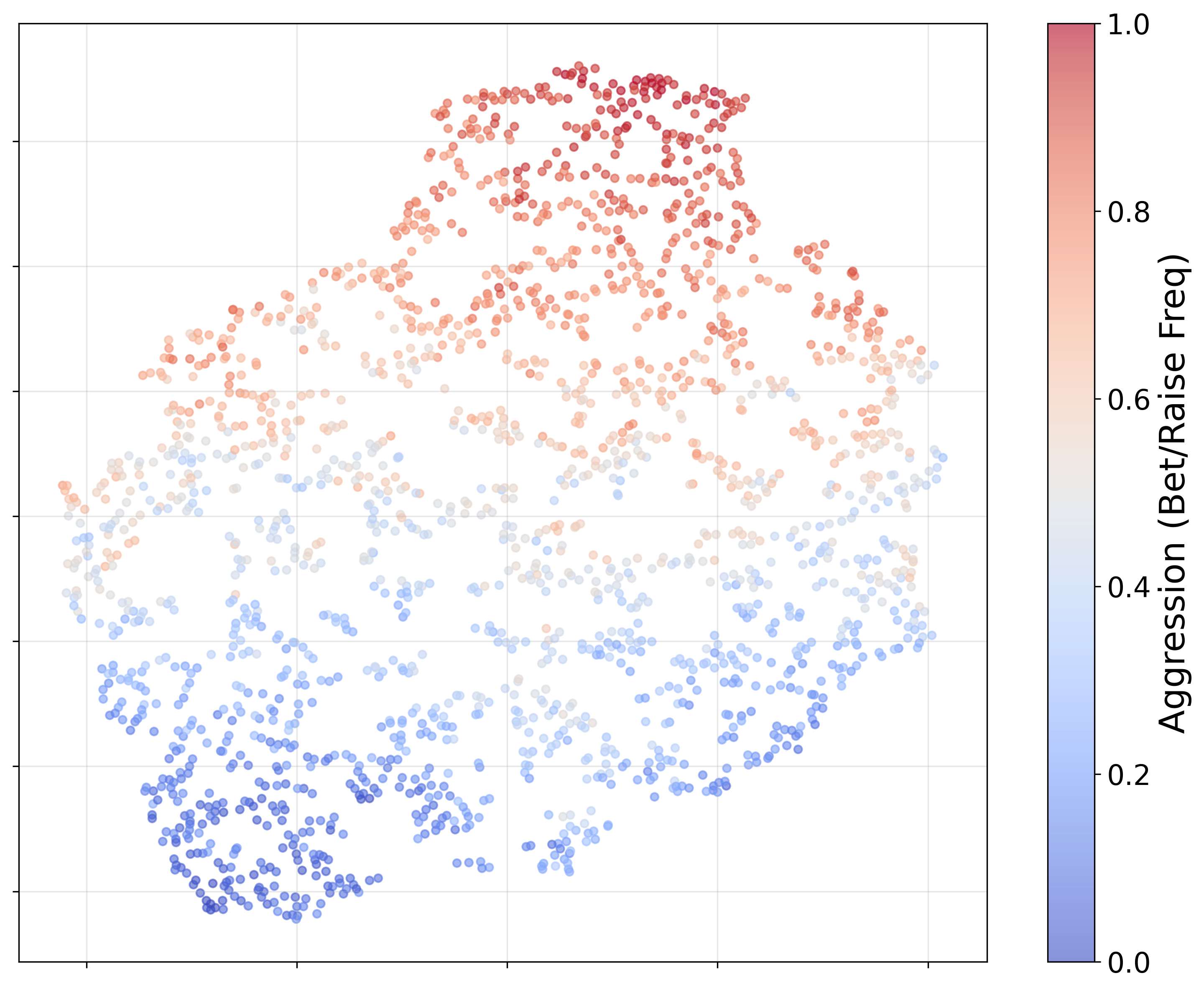}
        \footnotesize{(a) Trajectory-encoder embeddings} 
    \end{minipage}
    \hfill
    \begin{minipage}{0.38\textwidth}
        \centering
        \includegraphics[width=\linewidth]{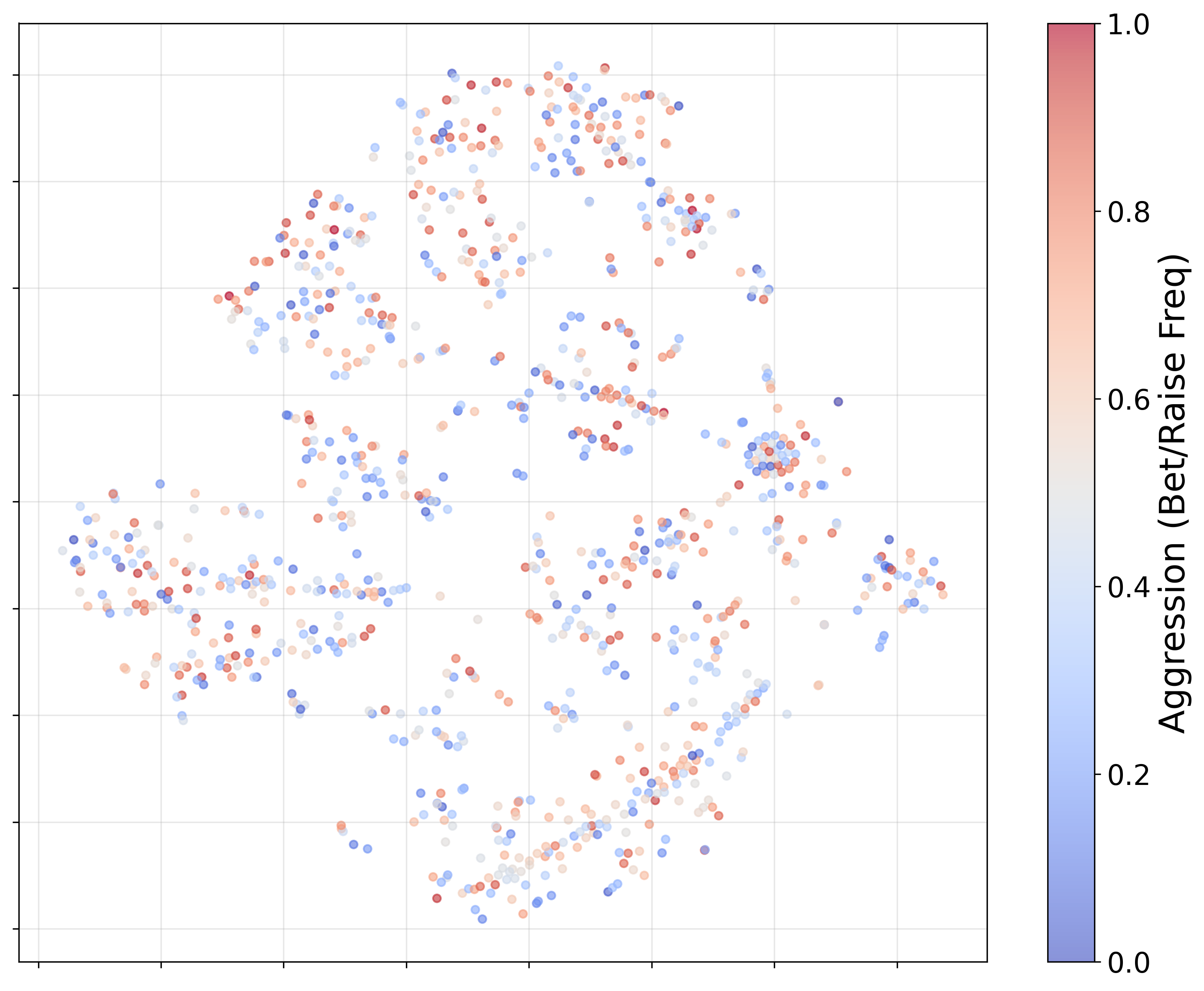}
        \footnotesize{(b) Weight autoencoder embeddings} 
    \end{minipage}

    \vspace{0.5em} 

    \caption{t-SNE visualization of embeddings in Kuhn Poker for 1000 random agents, then colored by aggression level (bet/raise frequency). With the trajectory encoder, aggressive strategies (red, top) clearly separated from passive strategies (blue, bottom).
    With the weight autoencoder, no such patterns emerge.}
    \label{fig:tsne_combined}
\end{figure*}

\subsection{Downstream Tasks}
\label{subsec:downstream-tasks}

In this section, we introduce a suite of downstream tasks to evaluate the usefulness of policy representations. Detailed descriptions in Appendix~\ref{app:downstream-tasks}.

\textbf{Task A (Payoff Prediction vs. Uniform Random).} How well can a linear predictor, given the embedding of a policy, predict the policy's expected payoff against a uniform random opponent?

\textbf{Task B (Payoff Prediction: Policy vs. Policy).} How well can a linear predictor, given the embeddings of a player 1 policy and a player 2 policy, predict the expected payoff of the two policies?

\textbf{Task C (Exploitability Prediction).} How well can a linear predictor, given the embedding of a policy, predict the payoff of an opponent's best-response to that policy? We compute ground-truth best-response values exactly, which is feasible since the games we use are small enough.

\textbf{Task D (zero-shot best-response)}. For a given Player 1 policy $\policy[1]$ with embedding $\embed[1]$, we train a zero-shot best-responder which, given $\embed[1]$, approximates a best-response against $\policy[1]$. The zero-shot best-responder is itself a NeuPL-style conditional policy neural network (Figure~\ref{fig:neupl}).

\textbf{Task E (Agent Identification).} Following \citet{grover2018learning}, we evaluate whether the embeddings are discriminative enough to identify which policy generated a given trajectory. For each of $N$ policies, we collect multiple held-out trajectories and embed them independently. A classifier (k-nearest neighbors or linear probe) is trained to predict the policy ID from the embedding.

\section{Experimental Results}

We test our methods on two zero-sum imperfect-information games: Kuhn Poker (12 information states) and Leduc Poker (936).
In Appendix~\ref{app:larger-games} we report preliminary results on two larger games: Liar's Dice (1 die, 4 sides; ${\sim}$1024), and Phantom Tic-Tac-Toe (${\sim}$500K). Rules for all games can be found in Appendix~\ref{sec:rules}.

\begin{table*}[t!]
    \centering
    \resizebox{0.95\textwidth}{!}{%
    \begin{tabular}{ll ccc c ccc}
    \toprule
    & & \multicolumn{3}{c}{\textbf{Kuhn Poker}} & & \multicolumn{3}{c}{\textbf{Leduc Poker}} \\
    \cmidrule(lr){3-5} \cmidrule(lr){7-9}
    \textbf{Task} & \textbf{Encoder} & \textbf{MSE} & \textbf{Baseline} & \textbf{Imp. (\%)} & & \textbf{MSE} & \textbf{Baseline} & \textbf{Imp. (\%)} \\
    \midrule
    {\textbf{Task A (Fixed)}}
    & Identity   & 0.097 $\pm$0.006 & 0.083 $\pm$0.005 & -16.9\% & & 0.374 $\pm$0.042 & 0.339 $\pm$0.050 & -10.3\% \\
    & Weight     & 0.081 $\pm$0.015 & 0.081 $\pm$0.016 & 0.0\% & & 0.350 $\pm$0.096 & 0.350 $\pm$0.091 & 0.0\% \\
    & Functional & 0.080 $\pm$0.010 & 0.080 $\pm$0.010 & 0.0\% & & 0.462 $\pm$0.009 & 0.462 $\pm$0.009 & 0.0\% \\
    & NeuPL      & 0.0208 $\pm$0.0022 & 0.0221 $\pm$0.0029 & 5.7\% & & 0.167 $\pm$0.037 & 0.196 $\pm$0.040 & 14.6\% \\
    & Tabular    & 0.022 $\pm$0.003 & 0.081 $\pm$0.016 & 71.9\% & & 0.266 $\pm$0.037 & 0.350 $\pm$0.091 & 22.0\% \\
    & Trajectory & 0.024 $\pm$0.001 & 0.083 $\pm$0.012 & 71.1\% & & 0.243 $\pm$0.047 & 0.336 $\pm$0.035 & 27.8\% \\
    & Grover     & 0.023 $\pm$0.001 & 0.083 $\pm$0.012 & 72.4\% & & 0.227 $\pm$0.031 & 0.336 $\pm$0.035 & 32.3\% \\
    \midrule
    {\textbf{Task B (Pair)}}
    & Identity   & 0.161 $\pm$0.017 & 0.151 $\pm$0.008 & -6.9\% & & 0.793 $\pm$0.131 & 0.764 $\pm$0.105 & -3.6\% \\
    & Weight     & 0.176 $\pm$0.004 & 0.160 $\pm$0.003 & -10.0\% & & 0.745 $\pm$0.093 & 0.709 $\pm$0.092 & -5.1\% \\
    & Functional & 0.135 $\pm$0.007 & 0.135 $\pm$0.006 & 0.0\% & & 0.738 $\pm$0.079 & 0.740 $\pm$0.080 & 0.3\% \\
    & Tabular    & 0.035 $\pm$0.002 & 0.160 $\pm$0.003 & 78.0\% & & \multicolumn{3}{c}{OOM ($>$32GB)} \\
    & Trajectory & 0.084 $\pm$0.005 & 0.160 $\pm$0.003 & 47.2\% & & 0.637 $\pm$0.006 & 0.704 $\pm$0.089 & 8.6\% \\
    & Grover     & 0.077 $\pm$0.006 & 0.160 $\pm$0.003 & 51.6\% & & 0.569 $\pm$0.020 & 0.705 $\pm$0.089 & 18.7\% \\
    \midrule
    {\textbf{Task C}}
    & Identity   & 0.304 $\pm$0.007 & 0.052 $\pm$0.002 & -485.7\% &  & 10.559 $\pm$0.326 & 1.149 $\pm$0.147 & -818.9\% \\
    & Weight     & 0.058 $\pm$0.010 & 0.056 $\pm$0.007 & -4.5\% &  & 1.511 $\pm$0.086 & 1.440 $\pm$0.006 & -4.9\% \\
    & NeuPL      & 0.036 $\pm$0.007 & 0.045 $\pm$0.013 & 19.3\% &  & 0.007 $\pm$0.001 & 0.035 $\pm$0.004 & 79.5\% \\
    \bottomrule
    \end{tabular}
    }
    \caption{Performance on payoff prediction.
    Each method was evaluated on randomly initialized policies, except NeuPL, which used NeuPL policies. Identity Task B uses 10K sampled train pairs and 2.5K sampled validation pairs due to the computational cost of evaluating all pairwise payoffs. \textit{Imp.} (improvement) denotes reduction in MSE from baseline to the method. Negative improvement occurs when the method does \textit{worse} than the baseline (e.g. due to overfitting).
    $\pm$ denotes std over 3 random seeds.
    }
    \label{tab:ab}
\end{table*}

\begin{table*}[t]
\centering
\resizebox{0.95\textwidth}{!}{%
\begin{tabular}{ll cccc c cccc}
\toprule
& & \multicolumn{4}{c}{\textbf{Kuhn Poker}} & & \multicolumn{4}{c}{\textbf{Leduc Poker}} \\
\cmidrule(lr){3-6} \cmidrule(lr){8-11}
\textbf{Encoder} & \textbf{Task} & \textbf{Payoff} & \textbf{Baseline} & \textbf{Imp.} & \textbf{Upp.} & & \textbf{Payoff} & \textbf{Baseline} & \textbf{Imp.} & \textbf{Upp.} \\
\midrule
{\textbf{Identity}}
& Task D & 0.253 $\pm$0.041 & 0.256 $\pm$0.092 & -0.003 & 0.497 $\pm$0.015 &  & -- & -- & -- & -- \\
\midrule
{\textbf{Weight}}
& Task D & 0.472 $\pm$0.029 & 0.412 $\pm$0.127 & +0.060 & 0.485 $\pm$0.034 &  & 2.124 $\pm$0.153 & 2.292 $\pm$0.099 & -0.168 & 3.172 $\pm$0.182 \\
\midrule

%
{\textbf{NeuPL}}
& Task D & 0.085 $\pm$0.012 & 0.038 $\pm$0.043 & +0.047 & 0.179 $\pm$0.010 & & 0.260 $\pm$0.022 & 0.080 $\pm$0.113 & +0.180 & 0.924 $\pm$0.015 \\
\bottomrule
\end{tabular}
}
\caption{
Performance on Task D. \textit{Payoff} is the empirical payoff of the best-responder to embedded policies. \textit{Baseline} is the empirical payoff of the best-responder when it is trained with random embeddings. \textit{Imp.} denotes the improvement from Baseline to Payoff (positive improvement is good). \textit{Upp.} denotes the upper bound: the average of the true best-response value for each embedded policy.
$\pm$ denotes std over 3 random seeds.
}
\label{tab:best-response-results}
\end{table*}

We evaluate the quality of our learned representations by training a linear predictor on the frozen embeddings for the downstream payoff prediction tasks defined in Section 4. As a \textbf{baseline} predictor for each task, we use the mean target across the training set. A qualitative visualization of the latent space is presented in \Cref{fig:tsne_combined}.

\paragraph{Tasks A and B.} We evaluate the identity baseline, the weight encoders, and NeuPL embeddings on the payoff prediction tasks (A and B). The results are in Table~\ref{tab:ab}. Neither the identity nor the weight encoders improve over simply predicting the mean.


In Kuhn Poker, where the tabular vector is compact and complete, it dominates on Task B.
On Task A, all methods perform comparably ($\sim$71--72\%). This is expected: a lossless 24-dimensional representation of the full policy is hard to beat in a game this small. However, as game complexity grows, the picture reverses. In Leduc, the tabular representation balloons to 2808 dimensions and \emph{underperforms} the learned 128-dimensional encoders on Task A (22.0\% vs.\ 27.8--32.3\%).

\paragraph{Tasks C and D.} In Task C, we embed Player 1 policies, and learn to predict the payoffs of Player 2's best-response to each Player 1 policy. Neither the identity baseline nor the weight reconstruction autoencoder perform better than predicting the mean on the dataset of random policies. NeuPL does improve over the baseline on the dataset of NeuPL policies.

In Task D~\ref{tab:best-response-results}, we similarly embed Player 1 policies and train a Player 2 zero-shot best-responder with PPO to best-respond to any given embedding. NeuPL again outperforms the naive baseline.\footnote{It is difficult to compare NeuPL to the other methods, since the population of policies used for training and evaluation is different than for the other methods.}



\begin{table}[th]
\centering
\begin{tabular}{ll cc}
\toprule
& & \textbf{Kuhn Poker} & \textbf{Leduc Poker} \\
\textbf{Encoder} & \textbf{Classifier} & \textbf{Top-1 (\%)} & \textbf{Top-1 (\%)} \\
\midrule
Trajectory & k-NN   & \textbf{39.9 $\pm$ 1.3} & \textbf{45.3 $\pm$ 1.8} \\
Trajectory & Linear & \textbf{47.9 $\pm$ 0.6} & \textbf{57.8 $\pm$ 0.9} \\
Grover     & k-NN   & 24.8 $\pm$ 0.1 & 34.0 $\pm$ 2.0 \\
Grover     & Linear & 25.7 $\pm$ 0.5 & 38.9 $\pm$ 0.7 \\
\bottomrule
\end{tabular}
\caption{Task E: Agent identification top-1 accuracy for 500 eval agents using 5 train / 5 test trajectories per agent. $\pm$ denotes std over 3 random seeds. Random baseline is 0.2\%.}
\label{tab:agent_id-small}
\end{table}

\paragraph{Trajectory Encoder vs. Grover.}
We compare against the Grover baseline \cite{grover2018learning}, the most closely related prior work, which learns embeddings from episode data using an MLP encoder with a hybrid generative-discriminative objective (imitation + triplet identification). Compared to our trajectory encoder, the two approaches show complementary strengths. On payoff prediction (\Cref{tab:ab}), Grover's hybrid objective yields a consistent advantage on Task A.
However, on Task E agent identification, our contrastive Transformer encoder substantially outperforms Grover,
achieving 47.9\% vs.\ 25.7\% top-1 accuracy on Kuhn and 57.8\% vs.\ 38.9\% on Leduc with a linear classifier (Table~\ref{tab:agent_id-small}).\footnote{Results on larger games in Appendix \ref{sec:task_x}}



\section{Conclusion}
In this work, we presented a comparative study of techniques and downstream tasks for learning policy representations in imperfect-information games.


The NeuPL-style embeddings show promise, and future work includes adding additional loss terms to further shape the embeddings.

Natural extensions to the downstream task suite include tasks involving belief-state prediction or integration with decision-time planning.

\section{Acknowledgments}
Thanks to Randall Balestriero for inspiring this work and giving helpful suggestions and discussions on it.

This material is based upon work supported by the National Science Foundation CISE Graduate Fellowships under Grant No. 2313998.

\FloatBarrier
\bibliography{bib}
\balance

\newpage
\appendix

\section{Details on Randomly Initializing Policy Neural Networks}
\label{sec:detail-init}

Our layer initialization initializes weights orthogonally with a scaling factor of $2.2$. It initializes the biases to $0$, except for the last layer (whose outputs are the action logits), for which each bias is randomly initialized between $-1$ and $1$.

This is in contrast to PyTorch's default layer initialization for linear layers, where weights are initialized according to a Kaiming uniform distribution, and biases are initialized randomly between $\sqrt{-b}, \sqrt{b}$, where $b$ is the fan-in of the neuron.

For the results in Table~\ref{tab:random-init-diversity-1}, we used 1000 policies randomly initialized via PyTorch default, 1000 policies randomly initialized with our method, 1000 policies generated by interpolating NeuPL embeddings, and 611 policies generated by running PSRO multiple times.

We train NeuPL with a population size of 25 per player. To generate each of the 1000 random policies, we randomly pick 2 of the agents, and interpolate uniformly at random to a convex combination of their embeddings.

\begin{table}[h]
    \centering
    \begin{tabular}{lcc}
        \toprule
        Initialization & Payoff & $P(\text{check} \mid J)$ \\
        \midrule
        PyTorch default & $0.124 \pm 0.022$ & $0.500 \pm 0.023$ \\
        Our init & $0.081 \pm \bf{0.258}$ & $0.505 \pm \bf{0.311}$ \\
        PSRO & $0.253 \pm 0.096$ & $0.386 \pm 0.297$ \\
        NeuPL & $0.145 \pm 0.096$ & $0.832 \pm 0.247$ \\
        \bottomrule
    \end{tabular}
    \caption{Policies with parameters randomly initialized with PyTorch's default, with our random initialization, generated via PSRO, and generated via NeuPL. Values are: payoffs vs. a uniform random strategy in Kuhn poker, and probability of taking the ``check'' action as the first player in Kuhn poker with a jack. Values are mean $\pm$ standard deviation. \textbf{Policies initialized with the default method are less diverse (they have much lower standard deviation in both values)}}
    \label{tab:random-init-diversity-1}
\end{table}

\section{Baseline Encoder Diagrams}
\label{sec:baseline-diagrams}

\begin{figure}[h!]
    \centering
    \includegraphics[width=0.65\linewidth]{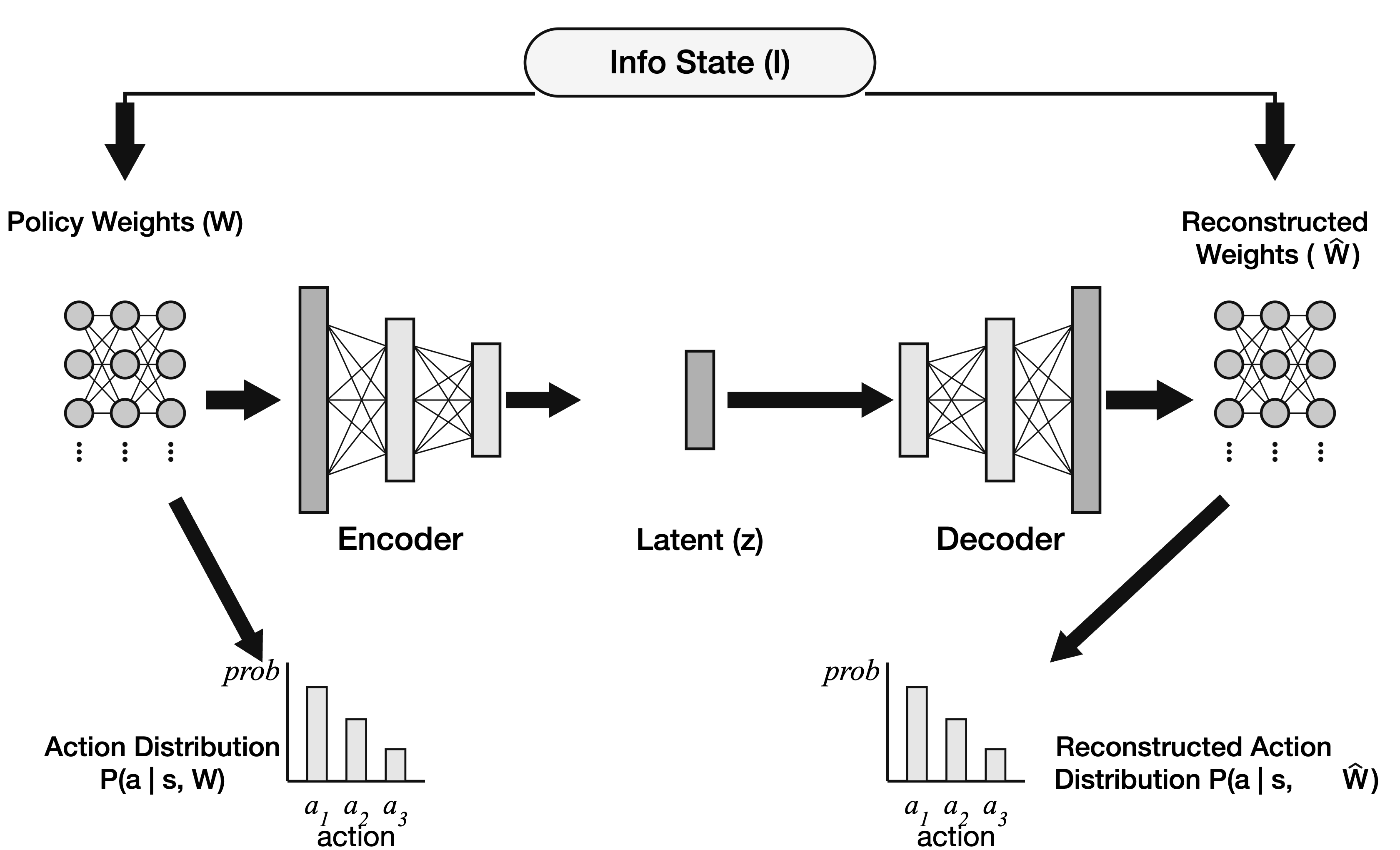}
    \caption{Functional encoder}
    \label{fig:fae}
\end{figure}

Illustration of functional encoder is in Fig~\ref{fig:fae}.

\section{Future Work}

While our experiments span games from ${\sim}$12 to ${\sim}$500K information states, scaling to even larger games such as 3x3 dark hex (millions of information states) and games with continuous action spaces remains an important direction.

Our opponent-adaptive strategy selection results highlight a key gap: representations trained for behavioral discrimination do not encode payoff-structural similarity in a linearly recoverable form, as demonstrated by our Mahalanobis projection experiment. A natural next step is to define training objectives that explicitly optimize for payoff-relevant structure, for example contrastive losses where positive pairs are agents with correlated payoff profiles rather than same-agent trajectories, or auxiliary losses that directly predict pairwise payoffs from embedding distances.

Finally, we are interested in downstream tasks that directly enable the creation of good policies in games, either via payoff prediction or as part of a decision-time planning process.

\section{Learned Similarity Metric}
\label{sec:learned_sim}

To test whether a better similarity metric could close the opponent adaptation gap, we trained a \emph{learned Mahalanobis projection} on Liar's Dice: a linear map $W \in \mathbb{R}^{64 \times 128}$ optimized on library agents so that $\cos(Wz_i, Wz_j)$ predicts payoff correlation between agents $i$ and $j$. The projection achieves $r{=}0.70$ correlation with ground-truth payoff similarity during training, confirming that it learns a meaningful transformation. However, kernel regression with the learned similarity (0.347 $\pm$ 0.030) does not outperform the simpler temperature-scaled cosine (0.351 $\pm$ 0.024) for either encoder, indicating that the bottleneck is not the similarity metric but the information content of the embeddings themselves.

\section{PSRO Population Results}
\label{sec:psro_results}

We additionally evaluated encoders on populations generated by the Policy Space Response Oracle algorithm (PSRO) \cite{lanctot2017unified}. PSRO iteratively expands a population by training best-response policies via deep RL against Nash equilibrium mixtures of the current population. This produces strategically structured agents, unlike the random populations used in the main experiments.

Results on Kuhn Poker are shown in \Cref{tab:psro_results}. The Trajectory Encoder maintains its advantage over weight-space methods on PSRO populations, consistent with the random population findings. PSRO training on Leduc Poker did not converge reliably within our computational budget due to the cost of computing best responses in the larger game tree; investigating scalable population generation for larger games is an important direction for future work.

\begin{table}[h]
\centering
\caption{Downstream task performance on PSRO-generated populations (Kuhn Poker only). $\pm$ denotes std over 3 random seeds where available.}
\label{tab:psro_results}
\begin{tabular}{ll ccc}
\toprule
\textbf{Encoder} & \textbf{Task} & \textbf{MSE} & \textbf{Baseline} & \textbf{Imp. (\%)} \\
\midrule
{\textbf{Identity}}
& Task A & 0.096 $\pm$ 0.007 & 0.030 $\pm$ 0.003 & -217.6\% \\
& Task C & 0.077 $\pm$ 0.005 & 0.0031 $\pm$ 0.0002 & -2384.5\% \\
\midrule
{\textbf{Weight}}
& Task A & 0.030 $\pm$ 0.003 & 0.029 $\pm$ 0.002 & -1.4\% \\
& Task C & 0.003609 $\pm$ 0.000007 & 0.003276 $\pm$ 0.000268 & -10.2\% \\
\midrule
{\textbf{Trajectory}}
& Task A & \textbf{0.0191} & 0.0240 & \textbf{20.4\%} \\
& Task B & \textbf{0.0281} & 0.0348 & \textbf{19.4\%} \\
\bottomrule
\end{tabular}
\end{table}



\section{Additional Results}

\subsection{Preliminary results on larger games}
\label{app:larger-games}
We performed preliminary experiments on the larger games of Liar's Dice and Phantom Tic-Tac-Toe. Results are reported in Table~\ref{tab:downstream_results_wide}

\begin{table*}[t!]
\centering
\resizebox{\textwidth}{!}{%
\begin{tabular}{ll ccc c ccc c ccc c ccc}
\toprule
& & \multicolumn{3}{c}{\textbf{Kuhn Poker}} & & \multicolumn{3}{c}{\textbf{Leduc Poker}} & & \multicolumn{3}{c}{\textbf{Liar's Dice}} & & \multicolumn{3}{c}{\textbf{Phantom TTT}} \\
\cmidrule(lr){3-5} \cmidrule(lr){7-9} \cmidrule(lr){11-13} \cmidrule(lr){15-17}
\textbf{Encoder} & \textbf{Task} & \textbf{MSE} & \textbf{Baseline} & \textbf{Imp. (\%)} & & \textbf{MSE} & \textbf{Baseline} & \textbf{Imp. (\%)} & & \textbf{MSE} & \textbf{Baseline} & \textbf{Imp. (\%)} & & \textbf{MSE} & \textbf{Baseline} & \textbf{Imp. (\%)} \\
\midrule
{\textbf{Tabular}}
& Task A (Fixed) & 0.022 $\pm$0.003 & 0.081 $\pm$0.016 & 71.9\% & & 0.266 $\pm$0.037 & 0.350 $\pm$0.091 & 22.0\% & & 0.016 $\pm$0.003 & 0.045 $\pm$0.008 & 63.7\% & & \multicolumn{3}{c}{OOM} \\
& Task B (Pair)  & \textbf{0.035 $\pm$0.002} & 0.160 $\pm$0.003 & \textbf{78.0\%} & & \multicolumn{3}{c}{OOM ($>$32GB)} & & \multicolumn{3}{c}{OOM ($>$32GB)} & & \multicolumn{3}{c}{OOM} \\
\midrule
{\textbf{Trajectory}}
& Task A (Fixed) & 0.024 $\pm$0.001 & 0.083 $\pm$0.012 & 71.1\% & & 0.243 $\pm$0.047 & 0.336 $\pm$0.035 & 27.8\% & & 0.019 $\pm$0.002 & 0.045 $\pm$0.008 & 56.6\% & & 0.019 $\pm$0.002 & 0.021 $\pm$0.002 & 9.0\% \\
& Task B (Pair)  & 0.084 $\pm$0.005 & 0.160 $\pm$0.003 & 47.2\% & & 0.637 $\pm$0.006 & 0.704 $\pm$0.089 & 8.6\% & & 0.047 $\pm$0.007 & 0.081 $\pm$0.005 & 41.9\% & & 0.025 $\pm$0.001 & 0.044 $\pm$0.002 & 43.0\% \\
\midrule
{\textbf{Grover}}
& Task A (Fixed) & 0.023 $\pm$0.001 & 0.083 $\pm$0.012 & 72.4\% & & \textbf{0.227 $\pm$0.031} & 0.336 $\pm$0.035 & \textbf{32.3\%} & & \textbf{0.015 $\pm$0.002} & 0.045 $\pm$0.008 & \textbf{67.2\%} & & \textbf{0.011 $\pm$0.001} & 0.021 $\pm$0.002 & \textbf{48.5\%} \\
& Task B (Pair)  & 0.077 $\pm$0.006 & 0.160 $\pm$0.003 & 51.6\% & & \textbf{0.569 $\pm$0.020} & 0.705 $\pm$0.089 & \textbf{18.7\%} & & \textbf{0.041 $\pm$0.005} & 0.081 $\pm$0.005 & \textbf{49.6\%} & & \textbf{0.024 $\pm$0.001} & 0.044 $\pm$0.002 & \textbf{45.3\%} \\
\bottomrule
\end{tabular}
}
\caption{Payoff prediction performance for encoders evaluated on the same shared agent pools. The Tabular baseline uses the complete action distribution at every information state as a fixed-dimensional representation (24-dim for Kuhn, 2808-dim for Leduc, 9216-dim for Liar's Dice). In Kuhn, the lossless tabular representation dominates on Task B. As game complexity grows, learned encoders overtake the tabular baseline despite compressing into 128 dimensions, and the tabular Task B exceeds 32GB memory (OOM). $\pm$ denotes std over 3 random seeds.}
\label{tab:downstream_results_wide}
\end{table*}

\subsection{Task E: Agent Identification}
\label{sec:task_x}

Following \citet{grover2018learning}, we evaluate whether the trajectory encoder embeddings can identify which policy generated a held-out trajectory. For 500 random agents, we collect 5 train and 5 test trajectories each, embed them independently, and train classifiers on the embeddings.

\begin{table}[h!]
\centering
\caption{Task E: Agent identification top-1 accuracy for 500 eval agents using 5 train / 5 test trajectories per agent. $\pm$ denotes std over 3 random seeds. Random baseline is 0.2\%.}
\label{tab:agent_id-full}
\resizebox{\columnwidth}{!}{%
\begin{tabular}{ll cccc}
\toprule
& & \textbf{Kuhn Poker} & \textbf{Leduc Poker} & \textbf{Liar's Dice} & \textbf{Phantom TTT} \\
\textbf{Encoder} & \textbf{Classifier} & \textbf{Top-1 (\%)} & \textbf{Top-1 (\%)} & \textbf{Top-1 (\%)} & \textbf{Top-1 (\%)} \\
\midrule
Trajectory & k-NN   & \textbf{39.9 $\pm$ 1.3} & \textbf{45.3 $\pm$ 1.8} & \textbf{83.8 $\pm$ 0.7} & 96.4 $\pm$ 0.5 \\
Trajectory & Linear & \textbf{47.9 $\pm$ 0.6} & \textbf{57.8 $\pm$ 0.9} & \textbf{89.5 $\pm$ 0.6} & \textbf{97.5 $\pm$ 0.3} \\
Grover     & k-NN   & 24.8 $\pm$ 0.1 & 34.0 $\pm$ 2.0 & 79.6 $\pm$ 2.6 & 96.5 $\pm$ 0.2 \\
Grover     & Linear & 25.7 $\pm$ 0.5 & 38.9 $\pm$ 0.7 & 76.6 $\pm$ 2.5 & 95.8 $\pm$ 0.4 \\
\bottomrule
\end{tabular}
}
\end{table}




\subsection{Our Trajectory Encoder vs. Grover}

On the larger games (Liar's Dice and Phantom TTT), both encoders achieve near-perfect identification and retrieval, with diminishing gaps. On strategy classification, both encoders perform comparably. These results indicate that the contrastive objective produces more fine-grained, discriminative representations suited to identification tasks, while the hybrid imitation objective better captures the smooth payoff structure needed for regression. We hypothesize that this gap arises because Grover's imitation loss directly optimizes for encoding action distributions that determine payoffs whereas the contrastive NT-Xent objective optimizes for inter-policy discrimination on a normalized embedding sphere, discarding magnitude information that may be relevant for regression.
An ablation over training trajectories per policy (Appendix~\ref{sec:traj_ablation}) shows that the contrastive encoder is sample-efficient: downstream performance is stable across a 5$\times$ range (10 to 50 trajectories), indicating that only modest behavioral data per agent is needed during pre-training.

\paragraph{Additional Identification and Retrieval Tasks.}
To further characterize how the two training objectives shape the embedding space, we evaluate on three additional downstream tasks: few-shot (1-shot) agent identification (Task E), policy retrieval (Task F), and strategy classification (Task G). These tasks probe different aspects of representation quality, from fine-grained individual discrimination to coarse behavioral categorization. Results are shown in \Cref{tab:new_tasks}.

The contrastive Trajectory Encoder consistently outperforms the Grover baseline on identification and retrieval tasks across all four games, though the gap narrows substantially on larger games. On Task E (\Cref{tab:agent_id-full}), the contrastive encoder achieves 89.5\% top-1 accuracy on Liar's Dice (Linear), compared to 76.6\% for Grover. On Phantom TTT, both encoders achieve near-perfect identification ($>$95\%), with the contrastive encoder maintaining a slight edge (97.5\% vs.\ 95.8\% Linear). On policy retrieval (\Cref{tab:new_tasks}), both encoders achieve Recall@10 $>$99\% on Liar's Dice and Phantom TTT, indicating that the correct agent appears in the top 10 nearest neighbors virtually every time. The gap is particularly striking in the 1-shot setting on Kuhn and Leduc, where the contrastive encoder achieves 33.1\% and 35.1\% (k-NN) compared to 18.2\% and 17.2\% for Grover, though on Phantom TTT both converge (91.5\% vs.\ 91.1\%).

On strategy classification (Task G), both encoders perform comparably: 79.7\% vs.\ 75.3\% on Kuhn and 56.0\% vs.\ 56.3\% on Leduc. This suggests that coarse-grained strategic style (aggression level) is captured by both training objectives. The substantially higher accuracy on Kuhn compared to Leduc for both encoders reflects the structural simplicity of Kuhn Poker, where aggression (bet frequency) is a near-complete description of a policy's behavior due to the minimal game tree (3 cards, one betting round). In Leduc, with 6 cards, a community card, and two betting rounds, policies vary along many more behavioral dimensions, making a single aggression scalar a coarser summary and the classification task correspondingly harder.

Taken together, these results reveal that different SSL objectives produce representations with different epistemic content: the contrastive objective captures fine-grained identity (knowing \emph{who}), while the hybrid objective better captures payoff structure (knowing \emph{how much}), and both capture broad behavioral categories. This decomposition, which form of knowledge about an opponent each objective prioritizes, is consistent with their designs: NT-Xent maximizes separation between distinct policies, while the Grover hybrid loss optimizes for action-distribution fidelity.


\begin{table}[h!]
\centering
\caption{Additional downstream tasks comparing Trajectory and Grover encoders. Few-shot ID uses 1 train trajectory per agent (Top-1 accuracy). Retrieval reports Recall@$k$. Strategy classification reports accuracy over 5 aggression buckets (random baseline: 20\%). $\pm$ denotes std over 3 seeds. ``--'' indicates tasks not applicable (strategy classification requires a game-specific style label not defined for dice/board games).}
\label{tab:new_tasks}
\resizebox{\columnwidth}{!}{%
\begin{tabular}{ll cccc}
\toprule
& & \textbf{Kuhn Poker} & \textbf{Leduc Poker} & \textbf{Liar's Dice} & \textbf{Phantom TTT} \\
\textbf{Task} & \textbf{Encoder} & & & & \\
\midrule
\multirow{2}{*}{Few-shot ID (k-NN)}
& Trajectory & \textbf{33.1 $\pm$ 0.3} & \textbf{35.1 $\pm$ 1.0} & -- & 91.5 $\pm$ 1.3 \\
& Grover     & 18.2 $\pm$ 1.5 & 17.2 $\pm$ 1.8 & -- & 91.1 $\pm$ 0.8 \\
\midrule
\multirow{2}{*}{Few-shot ID (Linear)}
& Trajectory & \textbf{31.4 $\pm$ 0.7} & \textbf{34.0 $\pm$ 0.8} & -- & \textbf{89.9 $\pm$ 1.4} \\
& Grover     & 16.2 $\pm$ 0.6 & 21.8 $\pm$ 0.7 & -- & 88.9 $\pm$ 0.7 \\
\midrule
\multirow{2}{*}{Retrieval R@1}
& Trajectory & \textbf{39.3 $\pm$ 0.5} & \textbf{45.3 $\pm$ 0.3} & \textbf{83.7 $\pm$ 1.2} & 96.7 $\pm$ 0.8 \\
& Grover     & 25.0 $\pm$ 0.8 & 30.7 $\pm$ 2.4 & 79.5 $\pm$ 1.0 & 96.5 $\pm$ 0.4 \\
\midrule
\multirow{2}{*}{Retrieval R@5}
& Trajectory & \textbf{75.8 $\pm$ 1.1} & \textbf{78.1 $\pm$ 0.5} & \textbf{97.8 $\pm$ 0.2} & 99.8 $\pm$ 0.1 \\
& Grover     & 56.7 $\pm$ 0.5 & 57.6 $\pm$ 2.5 & 96.2 $\pm$ 0.5 & 99.6 $\pm$ 0.1 \\
\midrule
\multirow{2}{*}{Retrieval R@10}
& Trajectory & \textbf{87.9 $\pm$ 0.2} & \textbf{87.9 $\pm$ 0.6} & \textbf{99.4 $\pm$ 0.1} & 100.0 $\pm$ 0.1 \\
& Grover     & 70.4 $\pm$ 1.0 & 70.2 $\pm$ 1.3 & 98.7 $\pm$ 0.4 & 99.9 $\pm$ 0.1 \\
\midrule
\multirow{2}{*}{Strategy Class.}
& Trajectory & \textbf{79.7 $\pm$ 3.8} & 56.0 $\pm$ 2.0 & -- & -- \\
& Grover     & 75.3 $\pm$ 1.2 & 56.3 $\pm$ 5.9 & -- & -- \\
\bottomrule
\end{tabular}
}
\end{table}

\paragraph{Opponent-Adaptive Strategy Selection.}
To evaluate whether the learned embeddings have practical utility beyond probe-based evaluation, we design a downstream application: embedding-based opponent-adaptive strategy selection. Given a novel opponent, the task is to select the best counter-strategy from a library of agents with known pairwise payoffs.

We split the 500 evaluation agents into 50 held-out test opponents and a 450-agent response library. We use \emph{kernel regression} to predict each library agent's payoff against the test opponent as a similarity-weighted combination of its known payoffs against all library agents, then select the agent with the highest predicted payoff. We compare against three baselines: \emph{Oracle} (best library agent per test opponent using ground-truth payoffs), \emph{Random} (uniformly random library agent), and \emph{Best-Avg} (library agent with highest mean payoff across the library, equivalent to kernel regression with uniform weights). Results on Liar's Dice and Leduc Poker are shown in \Cref{tab:opponent_adapt}.

\begin{table}[h!]
\centering
\caption{Opponent-adaptive strategy selection. Mean payoff from a 450-agent library against 50 held-out opponents. $\pm$ std over 10 seeds.}
\label{tab:opponent_adapt}
\resizebox{\columnwidth}{!}{%
\begin{tabular}{l cc cc}
\toprule
& \multicolumn{2}{c}{\textbf{Liar's Dice}} & \multicolumn{2}{c}{\textbf{Leduc Poker}} \\
\cmidrule(lr){2-3} \cmidrule(lr){4-5}
\textbf{Method} & \textbf{Traj.} & \textbf{Grov.} & \textbf{Traj.} & \textbf{Grov.} \\
\midrule
Oracle  & \multicolumn{2}{c}{0.538 $\pm$ .026} & \multicolumn{2}{c}{1.899 $\pm$ .077} \\
Best-Avg  & \multicolumn{2}{c}{0.343 $\pm$ .024} & \multicolumn{2}{c}{0.923 $\pm$ .157} \\
Random  & \multicolumn{2}{c}{$-$0.006 $\pm$ .030} & \multicolumn{2}{c}{$-$0.221 $\pm$ .056} \\
\midrule
Kernel ($\tau{=}1$)           & 0.348 $\pm$ .020 & 0.343 $\pm$ .025 & 0.903 $\pm$ .123 & 0.888 $\pm$ .151 \\
Kernel ($\tau{=}5$)           & \textbf{0.351 $\pm$ .024} & 0.344 $\pm$ .024 & \textbf{0.935 $\pm$ .132} & 0.897 $\pm$ .123 \\
\bottomrule
\end{tabular}%
}
\end{table}

On Liar's Dice, kernel regression with Trajectory embeddings and temperature scaling ($\tau{=}5$) achieves 0.351 vs.\ 0.343 for Best-Avg, winning on 9 of 10 seeds (sign test $p{\approx}0.02$). We observe the same pattern on Leduc Poker: Trajectory kernel ($\tau{=}5$) achieves 0.935 vs.\ 0.923 for Best-Avg, winning on 8 of 10 seeds, but the improvement is not statistically significant (paired $t$-test $p{=}0.68$). We additionally trained a learned Mahalanobis similarity metric on Liar's Dice, which did not outperform temperature-scaled cosine despite achieving $r{=}0.70$ correlation with ground-truth payoff similarity (\Cref{sec:learned_sim}), indicating that the bottleneck is the information content of the embeddings themselves.

\begin{figure}[h!]
\centering
    \includegraphics[width=\linewidth]{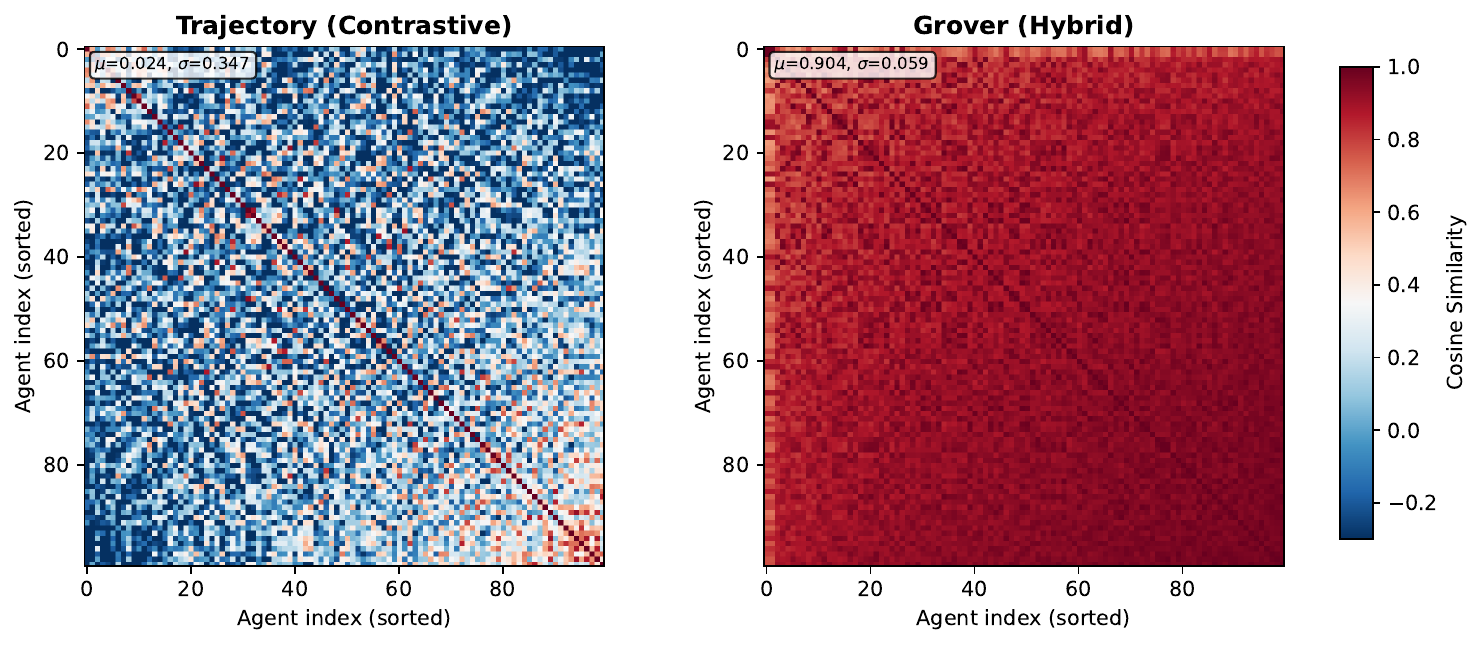}
    \caption{Cosine similarity matrices for 100 evaluation agents under Trajectory (contrastive) and Grover (hybrid) encoders on Liar's Dice. Agents are sorted by mean similarity. The Trajectory encoder produces structured similarity patterns, while the Grover encoder's similarities collapse to near-uniform values, explaining why its kernel regression degenerates to the Best-Avg baseline.}
    \label{fig:similarity_heatmap}
\end{figure}

The Grover encoder's kernel regression degenerates to Best-Avg on both games, consistent with its near-uniform cosine similarities ($\mu{=}0.90$, $\sigma{=}0.06$ on Liar's Dice; \Cref{fig:similarity_heatmap}). The large Oracle--Best-Avg gaps on both games confirm that opponent-specific information is exploitable in principle. These results, replicated across two games, multiple similarity metrics, and a supervised learned metric, reveal an epistemic blind spot in current SSL representations: they encode enough to recognize an opponent but not enough to know how to exploit them, and post-hoc correction via learned metrics cannot compensate for information that was never captured. Whether payoff-aware training objectives or uncertainty-aware representations that explicitly model what is \emph{not yet known} about an opponent could close this gap remains an open question.

\section{Training-Time Trajectory Ablation}
\label{sec:traj_ablation}

We ablate the number of trajectories per policy used during contrastive pre-training on Liar's Dice. \Cref{tab:traj_ablation} shows that downstream performance is remarkably stable across a 5$\times$ range of training trajectories (10 to 50 per policy). Even with only 10 trajectories per policy, the encoder achieves 55.5\% improvement on Task A, comparable to the default setting of 50 trajectories (55.6\%). This indicates that the contrastive encoder extracts discriminative policy representations efficiently, requiring only a modest number of behavioral observations per agent during pre-training.

\begin{table}[h]
\centering
\caption{Trajectory encoder ablation on Liar's Dice (seed 42): downstream payoff prediction as a function of training trajectories per policy.}
\label{tab:traj_ablation}
\begin{tabular}{c cc cc}
\toprule
\textbf{Traj./Policy} & \textbf{Task A MSE} & \textbf{Imp.} & \textbf{Task B MSE} & \textbf{Imp.} \\
\midrule
10  & 0.0169 & 55.5\% & 0.0447 & 43.7\% \\
25  & 0.0183 & 51.8\% & 0.0444 & 44.1\% \\
50  & 0.0168 & 55.6\% & 0.0458 & 42.4\% \\
\bottomrule
\end{tabular}
\end{table}

\section{Experimental Details and Hyperparameters}

\subsection{Code Implementation}

We use the OpenSpiel~\cite{lanctot_openspiel_2019} library. We extensively use a modified version of IIG-RL-Benchmark~\cite{rudolph2025reevaluating} for PPO training, PSRO, and NeuPL.

\subsection{Standard Training Configuration}
All experiments were conducted using three random seeds (42, 43, 44) and results were averaged, except for the opponent-adaptive strategy selection experiment which uses 10 seeds for statistical testing. For the Trajectory Encoder, we utilized 50 trajectories per policy during the training phase. Qualitative visualizations using t-SNE were generated with a perplexity value of 30.

\begin{table}[h]
\centering
\caption{Hyperparameters for Trajectory and Functional Encoders}
\begin{tabular}{l|cccc|cc}
\hline
\textbf{Parameter} & \textbf{Traj. (Kuhn)} & \textbf{Traj. (Leduc)} & \textbf{Traj. (LD)} & \textbf{Traj. (PTTT)} & \textbf{Func. (Kuhn)} & \textbf{Func. (Leduc)} \\ \hline
Num Agents & 500 & 500 & 500 & 500 & 500 & 500 \\
Batch Size & 16 & 16 & 16 & 8 & 16 & 16 \\
Epochs & 200 & 200 & 200 & 50 & 100 & 50 \\
Learning Rate & $1 \times 10^{-4}$ & $1 \times 10^{-4}$ & $1 \times 10^{-4}$ & $1 \times 10^{-4}$ & Default & Default \\
LR Scheduler & Cosine & Cosine & Cosine & Cosine & N/A & N/A \\
Val Split & 0.2 & 0.2 & 0.2 & 0.2 & N/A & N/A \\
Dataset Fraction & 1.0 & 1.0 & 1.0 & 1.0 & 1.0 & 0.02 \\ \hline
\end{tabular}

\vspace{1em}
The Grover encoder uses the same hyperparameters as described in the original work \cite{grover2018learning}: a 2-layer MLP encoder with 128-dimensional embeddings, trained with a hybrid imitation and triplet loss objective. All Grover models across the four games use the same architecture and training configuration.
\end{table}

\subsection{Hardware and Infrastructure}
Experiments were executed on an \textbf{NVIDIA Quadro RTX 6000} (24GB VRAM) and an \textbf{Intel(R) Xeon(R) Platinum 8268 CPU @ 2.90GHz}. For Identity Task B, we evaluate sampled agent pairs because computing payoffs for the full pairwise evaluation set was too expensive for our available compute budget.

\section{Reproducibility: CLI Commands}
To reproduce the specific checkpoints mentioned in the results, the following commands can be executed:

\begin{small}
\begin{verbatim}
# Trajectory Encoder (Kuhn Random)
python trajectory_encoder.py --game kuhn_poker --num-agents 500 --batch-size 16 \
--epochs 200 --lr 1e-4 --val-split 0.2 --lr-scheduler cosine --normalize

# Trajectory Encoder (Leduc Random)
python trajectory_encoder.py --game leduc_poker --num-agents 500 --batch-size 16 \
--epochs 200 --lr 1e-4 --val-split 0.2 --lr-scheduler cosine --normalize

# Functional Encoder (Kuhn Random)
python functional_autoencoder.py --game kuhn_poker --num-agents 500 --batch-size 16 \
--epochs 100

# Trajectory Encoder (Liar's Dice Random)
python trajectory_encoder.py --game "liars_dice(numdice=1,dice_sides=4)" \
--agent-pool agent_pools/liars_dice_numdice1_dice_sides4_seed42_n500.pt \
--num-agents 500 --batch-size 16 --epochs 200 --lr 1e-4 --val-split 0.2 \
--lr-scheduler cosine --normalize

# Trajectory Encoder (Phantom TTT Random)
python trajectory_encoder.py \
--game "phantom_ttt(obstype=reveal-nothing)" \
--agent-pool agent_pools/phantom_ttt_seed42_n500.pt \
--num-agents 500 --batch-size 8 --epochs 50 --lr 1e-4 --val-split 0.2 \
--lr-scheduler cosine --normalize

# Functional Encoder (Leduc Random)
python functional_autoencoder.py --game leduc_poker --num-agents 500 --batch-size 16 \
--epochs 50 --dataset-fraction 0.02 --device cuda
\end{verbatim}
\end{small}

\section{Detailed Performance Metrics: Random Agents}

The following table summarizes the Mean Squared Error (MSE) and corresponding baseline MSE for Tasks A and B using 1,000 random agents (500 for training/upstream and 500 for evaluation/downstream). Trajectory and Grover results use shared pre-generated agent pools to ensure identical evaluation populations across encoders. For models where three seeds were available, results represent individual run samples.

\begin{table}[H]
\centering
\caption{MSE Results per Seed for Random Agent Encoders}
\resizebox{\textwidth}{!}{%
\begin{tabular}{ll|ccc|ccc|ccc|ccc}
\hline
 & & \multicolumn{3}{c|}{\textbf{Kuhn}} & \multicolumn{3}{c|}{\textbf{Leduc}} & \multicolumn{3}{c|}{\textbf{Liar's Dice}} & \multicolumn{3}{c}{\textbf{Phantom TTT}} \\
\textbf{Encoder} & \textbf{Task} & \textbf{S42} & \textbf{S43} & \textbf{S44} & \textbf{S42} & \textbf{S43} & \textbf{S44} & \textbf{S42} & \textbf{S43} & \textbf{S44} & \textbf{S42} & \textbf{S43} & \textbf{S44} \\ \hline
Identity & Task A & 0.0961 & 0.0924 & 0.1036 & 0.3771 & 0.3298 & 0.4144 & -- & -- & -- & -- & -- & -- \\
 & Baseline & 0.0798 & 0.0807 & 0.0890 & 0.3476 & 0.2856 & 0.3849 & -- & -- & -- & -- & -- & -- \\
 & Task B & 0.1703 & 0.1415 & 0.1720 & 0.6970 & 0.7396 & 0.9427 & -- & -- & -- & -- & -- & -- \\
 & Baseline & 0.1549 & 0.1418 & 0.1550 & 0.6765 & 0.7342 & 0.8802 & -- & -- & -- & -- & -- & -- \\ \hline
Weight & Task A & 0.0640 & 0.0923 & 0.0860 & 0.3870 & 0.2409 & 0.4212 & -- & -- & -- & -- & -- & -- \\
 & Baseline & 0.0630 & 0.0940 & 0.0853 & 0.3828 & 0.2472 & 0.4191 & -- & -- & -- & -- & -- & -- \\
 & Task B & 0.1718 & 0.1777 & 0.1795 & 0.7159 & 0.6690 & 0.8487 & -- & -- & -- & -- & -- & -- \\
 & Baseline & 0.1612 & 0.1568 & 0.1622 & 0.6529 & 0.6590 & 0.8144 & -- & -- & -- & -- & -- & -- \\ \hline
Functional & Task A & 0.0688 & 0.0885 & 0.0841 & 0.4517 & 0.4697 & 0.4639 & -- & -- & -- & -- & -- & -- \\
 & Baseline & 0.0688 & 0.0885 & 0.0841 & 0.4517 & 0.4697 & 0.4633 & -- & -- & -- & -- & -- & -- \\
 & Task B & 0.1425 & 0.1347 & 0.1282 & 0.6670 & 0.8237 & 0.7248 & -- & -- & -- & -- & -- & -- \\
 & Baseline & 0.1410 & 0.1352 & 0.1292 & 0.6686 & 0.8277 & 0.7249 & -- & -- & -- & -- & -- & -- \\ \hline
Trajectory & Task A & 0.0240 & 0.0243 & 0.0225 & 0.2454 & 0.2887 & 0.1938 & 0.0168 & 0.0210 & 0.0206 & 0.0169 & 0.0208 & 0.0192 \\
 & Baseline & 0.0704 & 0.0934 & 0.0850 & 0.3046 & 0.3739 & 0.3285 & 0.0379 & 0.0541 & 0.0440 & 0.0194 & 0.0209 & 0.0223 \\
 & Task B & 0.0792 & 0.0853 & 0.0892 & 0.6434 & 0.6316 & 0.6368 & 0.0458 & 0.0409 & 0.0551 & 0.0261 & 0.0243 & 0.0247 \\
 & Baseline & 0.1624 & 0.1572 & 0.1610 & 0.6553 & 0.6518 & 0.8063 & 0.0794 & 0.0770 & 0.0868 & 0.0439 & 0.0423 & 0.0458 \\ \hline
Grover & Task A & 0.0236 & 0.0217 & 0.0223 & 0.2163 & 0.2629 & 0.2032 & 0.0125 & 0.0171 & 0.0149 & 0.0095 & 0.0116 & 0.0111 \\
 & Baseline & 0.0704 & 0.0934 & 0.0850 & 0.3046 & 0.3739 & 0.3285 & 0.0379 & 0.0541 & 0.0440 & 0.0194 & 0.0209 & 0.0223 \\
 & Task B & 0.0702 & 0.0824 & 0.0795 & 0.5686 & 0.5495 & 0.5894 & 0.0396 & 0.0367 & 0.0468 & 0.0243 & 0.0235 & 0.0243 \\
 & Baseline & 0.1623 & 0.1566 & 0.1615 & 0.6565 & 0.6518 & 0.8072 & 0.0794 & 0.0770 & 0.0868 & 0.0439 & 0.0423 & 0.0458 \\ \hline
Tabular & Task A & 0.0247 & 0.0210 & 0.0195 & 0.2800 & 0.2246 & 0.2938 & 0.0134 & 0.0198 & 0.0163 & OOM & OOM & OOM \\
 & Baseline & 0.0630 & 0.0940 & 0.0853 & 0.3828 & 0.2472 & 0.4191 & 0.0379 & 0.0541 & 0.0440 & -- & -- & -- \\
 & Task B & 0.0335 & 0.0349 & 0.0370 & OOM & OOM & OOM & OOM & OOM & OOM & OOM & OOM & OOM \\
 & Baseline & 0.1612 & 0.1568 & 0.1622 & -- & -- & -- & -- & -- & -- & -- & -- & -- \\ \hline
\end{tabular}
}
\end{table}

\subsection{Performance Observations}
\begin{itemize}
    \item \textbf{Identity Baseline}: The identity encoder (raw parameters) does not improve over the mean predictor on Task B, even when evaluated with sampled agent pairs due to the computational cost of the full pairwise evaluation.
    \item \textbf{Tabular Baseline}: The complete action distribution dominates on Kuhn Task B (MSE $\sim$0.035 vs.\ $\sim$0.077--0.084 for learned encoders), confirming that in small games, the lossless tabular policy is hard to beat. In Leduc and Liar's Dice, the high-dimensional tabular vectors (2808-dim, 9216-dim) underperform learned 128-dim encoders on Task A, and Task B exceeds 32GB memory (OOM) due to the high-dimensional embedding pairs. On Phantom TTT (${\sim}$500K information states), the tabular representation is entirely infeasible (OOM on both tasks).
    \item \textbf{Grover vs.\ Trajectory}: The Grover encoder outperforms the Trajectory encoder on Tasks A and B across all four games. The gap is most pronounced on Phantom TTT (48.5\% vs.\ 9.0\% on Task A) and Liar's Dice (67.2\% vs.\ 56.6\% on Task A, 49.6\% vs.\ 41.9\% on Task B).
    \item \textbf{Weight and Functional Stability}: Both the Weight Autoencoder and Functional Autoencoder largely mirrored baseline performance on random agents, suggesting that simple parameter or action distribution reconstruction is less effective than trajectory-based modeling for these tasks.
\end{itemize}

\section{Game Rules}
\label{sec:rules}
\paragraph{Kuhn Poker} is the simplest possible version of poker. It was developed by game theorist Harold Kuhn in 1950 to study simplified zero-sum games \cite{kuhn2016simplified}. It strips poker down to its absolute mathematical core: one card, one bet and two players (P1 and P2). The deck consists of 3 cards one King $K$, one Queen $Q$ and a Jack $J$. Each player enters a nominal ante into the pot before being dealt one card. There is one betting round. The first player can Check or Bet. If P1 checks, P2 can check (followed by an immediate showdown), or Bet. If P1 Bets, player 2 can fold (P1 wins) or call (showdown). If P2 bets (following P1 check), P1 can fold or call. If neither player folds, both players reveal their cards in the showdown where $K > Q > J$. The player with the highest card wins the pot. The game is fully solved. We know the exact Nash Equilibrium strategy (e.g., if holding a Jack, Player 1 should never bet, but if holding a King, they should bet 3 times as often as they bet with a Queen to balance their bluffing range).

\paragraph{Leduc Poker} adds a community card and multiple rounds of betting making it significantly more complicated and a better proxy for Texas Hold'em. In this game there are two players and 6 cards: two Kings, Queens and Jacks. To start, both players pay an ante and a then dealt one private card. \textbf{Round 1:} Players bet based on their private card. There is a fixed bet size. \textbf{The Flop:} One community card is revealed from the remaining deck. \textbf{Round 2:} Players bet again now knowing their card and the community card. \textbf{The Showdown:} Both players reveal their cards; A pair is a winning hand, otherwise the player with the highest card wins. If both players have the same card, the pot is split.

\paragraph{Liar's Dice} (1 die, 4 sides) is a bluffing game structurally distinct from poker. Each of the two players privately rolls a single four-sided die. Players alternate making claims about the total count of a particular face value across \emph{both} dice combined (e.g., ``there are at least two 3s''). Each claim must be strictly higher than the previous one (either a higher count or a higher face value). A player may instead challenge the opponent's last claim. If challenged, both dice are revealed: if the claim is true, the challenger loses; otherwise, the claimant loses. The game requires reasoning about hidden information (the opponent's die) and incentivizes bluffing where players can make inflated claims to pressure opponents into folding or to set up a trap. With 1 die and 4 sides, the game has approximately 1024 information states and 9 possible actions per state.

\paragraph{Phantom Tic-Tac-Toe} is a partially observable variant of Tic-Tac-Toe where players cannot see their opponent's moves. Each player places marks on a 3$\times$3 board, but only observes their own marks and whether an attempted move was rejected (because the cell was already occupied). If a player attempts to place on an occupied cell, they are informed the move is invalid and must choose again, but they do not learn who occupies that cell. The game ends when a player completes a row, column, or diagonal, or when the board is full (draw). With the ``reveal-nothing'' observation type, the game has approximately 500,000 information states, making it the largest game in our benchmark and a test of scalability beyond card games.

\section{Extended Related Works}
\label{sec:related-works}

\paragraph{Agent Modeling and Policy Representations}
Learning representations of other agents' behavior is a central challenge in multi-agent systems. \citet{albrecht_autonomous_2018} and \citet{nashed_survey_2022} performed comprehensive surveys on opponent modeling. Early work by \citet{he2016opponent} introduced DRON for opponent modeling via deep RL, and \citet{grover2018learning} learn low-dimensional policy embeddings using variational inference, evaluating on agent identification and reward prediction (our Task E follows their identification protocol). While \citet{grover2018learning} desire ``simulating the agent's policy'' and ``distinguishing the agent's policy'' as the two desiderata of good representations, we focus more on predicting payoffs in two-player zero-sum games as a measure of good representations. More recently, \citet{papoudakis2021agent} learn agent representations under partial observability by training an encoder-decoder to predict other agents' actions from local observations. \citet{xie2020learning} learn latent representations of other agents' strategies to influence multi-agent interactions. Most closely related to our trajectory encoder, CLAM \cite{ma2024clam} uses contrastive learning on local observations to produce real-time policy representations of other agents, and TransAM \cite{wallace2025transam} uses a transformer to encode local trajectories into embeddings that capture other agents' policies. Our work differs from these approaches in that we (i) focus specifically on imperfect-information games, (ii) systematically compare multiple types of policy embeddings and (iii) evaluate on game-theoretic downstream tasks such as payoff and exploitability prediction rather than cooperative returns.

\paragraph{Behavioral Diversity and Trajectory Representations}
Several works have used trajectory-level representations to measure or encourage policy diversity. TrajeDi \cite{lupu2021trajectory} defines behavioral distances between policies in trajectory space for zero-shot coordination. CURL \cite{laskin2020curl} applies contrastive learning to state observations for sample-efficient RL, though it learns state representations for a single agent rather than \emph{policy} representations across agents. Our trajectory encoder adapts contrastive learning to the latter setting, learning embeddings that are invariant to stochastic noise but discriminative of strategic intent.

\subsection{Grover}

Details on \citet{grover2018learning}:

The hybrid loss is defined as:

\begin{align*}
    \calL_{\text{Grover}} = \calL_{\text{imit}} + \lambda \, \calL_{\text{id}} \\
    \quad \calL_{\text{id}} = \big(1 + \exp(\|r_e - n_e\|_2 - \|r_e - p_e\|_2)\big)^{-2},
\end{align*}
where $p_e$, $r_e$, $n_e$ are embeddings of a positive, reference, and negative episode respectively.

\subsection{Additional Related Works}

\paragraph{Chess style learning} Other relevant research directions include learning individual players' styles in chess. This was done by \citet{mcilroy-young_learning_2022}. Their approach fine-tunes policy networks to represent styles of different players. The methods in this paper could be used to represent styles of different players instead as low-dimensional embeddings.

\paragraph{Weight-Space Learning}
Several works have explored learning from or about neural network weights directly. \citet{unterthiner2020predicting} showed that network properties can be predicted from weight statistics alone, while hypernetworks \cite{ha2016hypernetworks} generate weights for target networks. Our weight and functional autoencoders draw on this line of work, but our results suggest that weight-space proximity is a poor proxy for behavioral similarity in games.

\section{Details on Downstream Tasks}
\label{app:downstream-tasks}
\subsection{Task A}

A linear predictor receives a single agent embedding $\embed[1]$ and predicts its expected payoff against a static, uniform random opponent $\policy[\text{rand}]$. This tests whether the embeddings correlate with general strategic competence.

We want the linear prediction (left side of the equation) to approximate the expected returns (right side of the equation):
$$
\hat{y}_A = \mathbf{w}_A^\top \embed[1] + b_A \approx \E_{\policy[1], \policy[\text{rand}]} \bigg[\sum_{t=0}^{\tmax} \Rewardfunc(\state[][t], \action[][t])\bigg].
$$

\subsection{Task B}

The linear probe receives the concatenated embeddings of two agents and predicts the payoff of a head-to-head match:
$$
\hat{y}_B = \mathbf{w}_B^\top [\embed[1] ; \embed[2]] + b_B \approx \E_{\policy[1], \policy[2]} \bigg[\sum_{t=0}^{\tmax} \Rewardfunc(\state[][t], \action[][t])\bigg].
$$
This tests whether the latent space captures relational information about how specific strategies interact, not just their individual quality.

\subsection{Task C}

For a given Player 1 policy $\policy[1]$, the best-response value is defined as $\text{BRV}(\policy[1]) = \max_{\policy[2]} \E_{\policy[1], \policy[2]}[-\sum_t \Rewardfunc(\state[][t], \action[][t])]$. A linear probe predicts this from the embedding:
$$
\hat{y}_C = \mathbf{w}_C^\top \embed[1] + b_C \approx \text{BRV}(\policy[1]).
$$




\end{document}